\documentclass[journal]{IEEEtran}
\usepackage{ifpdf}
\usepackage{color}
\usepackage{amssymb}
\usepackage[linesnumbered,ruled,vlined]{algorithm2e}
\usepackage{cite}
\usepackage{amsmath}
\usepackage{units}
\usepackage{array}
\usepackage{fixltx2e}
\usepackage{stfloats}
\usepackage{url}
\usepackage{hyperref}
\usepackage{subfigure}
\usepackage{footnote}
\usepackage{multirow}
\usepackage{booktabs}
\usepackage{tabularx}

\ifCLASSINFOpdf
  \usepackage[pdftex]{graphicx}
\else
\fi

\begin{document}
\title{Towards Human-Centric Autonomous Driving: A Fast-Slow Architecture Integrating Large Language Model Guidance with Reinforcement Learning
}

\author{Chengkai~Xu,
        Jiaqi~Liu, 
        Yicheng~Guo, 
        Yuhang~Zhang,
        Peng~Hang,~\IEEEmembership{Senior Member,~IEEE,}
        and~Jian~Sun
\thanks{This work was supported in part by the National Natural Science Foundation of China (52302502), the State Key Laboratory of Intelligent Green Vehicle and Mobility under Project No. KFZ2408, the State Key Lab of Intelligent Transportation System under Project No. 2024-A002, and the Fundamental Research Funds for the Central Universities.}
\thanks{C. Xu, J. Liu, Y. Guo, Y. Zhang, P. Hang and J. Sun are with the College of Transportation, Tongji University, Shanghai 201804, China. (e-mail: xck1270157991@gmail.com, {liujiaqi13, guo\_yicheng, 2153338, hangpeng, sunjian}@tongji.edu.cn)}
\thanks{Corresponding author: Peng Hang}}


\maketitle

\begin{abstract}
Autonomous driving has made significant strides through data-driven techniques, achieving robust performance in standardized tasks. However, existing methods frequently overlook user-specific preferences, offering limited scope for interaction and adaptation with users.
To address these challenges, we propose a “fast-slow” decision-making framework that integrates a Large Language Model (LLM) for high-level instruction parsing with a Reinforcement Learning (RL) agent for low-level real-time decision. In this dual system, the LLM operates as the “slow” module, translating user directives into structured guidance, while the RL agent functions as the “fast” module, making time-critical maneuvers under stringent latency constraints. By decoupling high-level decision making from rapid control, our framework enables personalized user-centric operation while maintaining robust safety margins. Experimental evaluations across various driving scenarios demonstrate the effectiveness of our method. Compared to baseline algorithms, the proposed architecture not only reduces collision rates but also aligns driving behaviors more closely with user preferences, thereby achieving a human-centric mode. By integrating user guidance at the decision level and refining it with real-time control, our framework bridges the gap between individual passenger needs and the rigor required for safe, reliable driving in complex traffic environments.
\end{abstract}


\IEEEpeerreviewmaketitle

\section{Introduction}
\IEEEPARstart{I}{n} recent years, rapid advancements in hardware and machine learning technologies have spurred significant progress in autonomous driving, positioning it to reshape modern transportation systems\cite{chen2022milestones, zhou2024vision}. Despite these strides, a critical gap remains in achieving human-centric design, that is, the capacity of an autonomous system to interpret and accommodate diverse user preferences\cite{gebru2022review}, such as “please speed up, I’m running late”. Traditional data-driven or rule-based methods often struggle to translate such high-level, potentially ambiguous instructions into effective low-level control actions. Moreover, many existing approaches still lack a robust mechanism for integrating user feedback, limiting their ability to flexibly adapt to evolving demands in real-world settings\cite{kuznietsov2024explainable, chen2024end}.

Reinforcement Learning (RL), due to its excellent learning and generalization capabilities, has been widely used in the design of autonomous driving decision algorithms\cite{kiran2021deep, wu2024recent}. RL-based approaches have demonstrated strong performance in well-defined tasks\cite{haydari2020deep}. Yet, RL systems typically treat user commands as either non-existent or too simplistic, hindering their ability to capture nuanced or evolving human intentions. In contrast, rule-based methods can encode certain human guidelines more directly, but they struggle to adapt flexibly to novel or rare scenarios, leading to suboptimal or brittle performance.

Meanwhile, Large Language Models (LLMs), represented by Deepseek\cite{liu2024deepseek} and ChatGPT\cite{achiam2023gpt}, have achieved remarkable success in natural language understanding and generation. Their generative capabilities enable them to parse complex or abstract directives, potentially offering a powerful interface for human-vehicle interaction. However, despite their proficiency in language tasks, LLMs alone are not sufficient for real-time, safety-critical control. Moreover, without explicit constraints, they might produce instructions that overlook safety margins, traffic laws, or physical feasibility\cite{wen2023dilu}. These considerations make LLMs more suitable as a “slow” system\cite{sha2023languagempc}, where their interpretative strength can be harnessed to translate human preferences into structured context-aware guidance, while delegating rapid, fine-grained control decisions to an RL agent.

\begin{figure}
    \centering
    \includegraphics[width=0.9\linewidth]{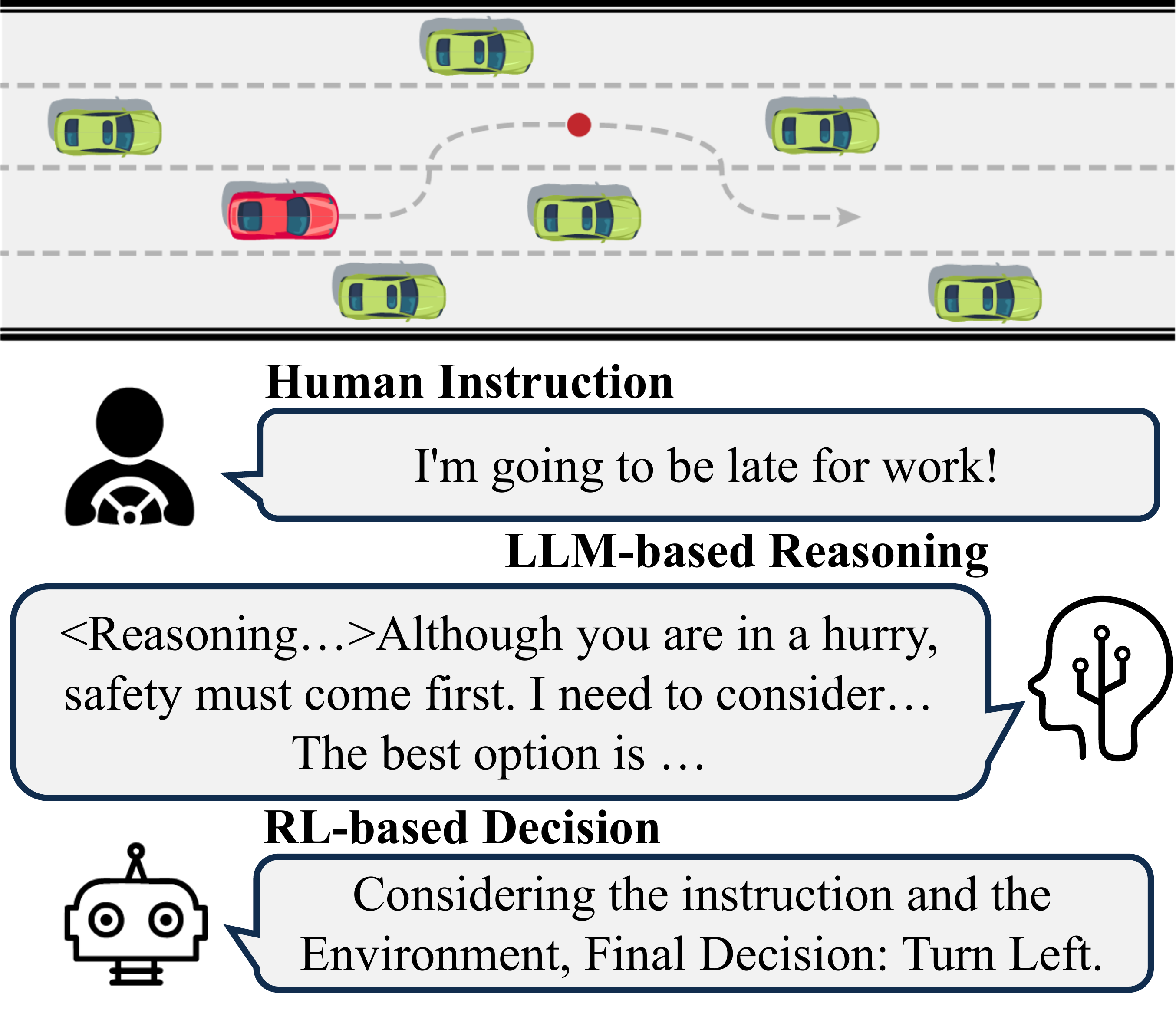}
    \caption{The LLM interprets high-level human intent and generates structured guidance, while the RL module integrates this guidance with environmental context to make optimal, human-aligned decisions.}
    \label{fig:overview}
\end{figure}

To bridge these complementary strengths, we propose a “fast-slow” decision-making framework that pairs an LLM for high-level command parsing with an RL agent for low-level vehicle control. As shown in Fig.~\ref{fig:overview}, the LLM processes user-provided commands and outputs structured guidance signals. The RL module, operating under real-time constraints, then adjusts steering and acceleration based on the LLM's commands, ensuring adaptability and robust safety. Using the LLM interpretation ability and RL efficiency, our framework addresses the gap between pure data-driven approaches that are unconstrained and knowledge-driven approaches that require real-time performance to be guaranteed. 

Our contributions can be summarized as follows:

\begin{itemize}
    \item A novel human-centric two-tier decision-making framework is proposed, which integrates a LLM for interpreting high-level user instructions with an RL agent for real-time, low-level decision-making.
    \item An adaptive coordination mechanism is designed, enabling the RL agent to selectively delay or override directives from the LLM when safety constraints demand, thereby balancing user preference and risk.
    \item Empirical evaluation of the proposed framework is conducted across a range of driving scenarios, demonstrating superior safety and greater adherence to user preferences compared to existing baselines.
\end{itemize}

\section{Related Works}
\subsection{Reinforcement Learning for Autonomous Driving}
RL methods have been increasingly applied in autonomous driving tasks such as lane-changing, adaptive cruise control, and overtaking maneuvers\cite{wu2024continuous, liu2025language, li2023overcoming}. While these data-driven methods frequently exhibit strong performance in simulated settings, several challenges remain.


Most RL agents are developed and evaluated with fixed objectives, unable to incorporate evolving user demands that might arise in real-world driving. Approaches that accept human input typically treat it as static or overly simplistic, limiting their responsiveness to dynamic user preferences\cite{kaufmann2023survey, wu2023toward, wu2024human}. Meanwhile, although RL excels in well-defined tasks with consistent reward structures, it remains susceptible to poor generalization under distributional shifts when encountering ambiguous instructions from passengers\cite{cobbe2019quantifying, hansen2021generalization}. These issues underscore the need for more adaptable frameworks that can reconcile robust policy learning with diverse user goals.

\subsection{Human-Centric Autonomous System}
In parallel to RL research, there has been growing attention to human-centric autonomous systems\cite{yang2024human, gebru2022review}. The main objective is to design methods that can intuitively interpret human preferences and incorporate them into the control loop. Early works mainly leverage voice or text-based interfaces, using rule-based natural language parsers or simpler machine learning models to allow limited customization of vehicle parameters\cite{chen2025human, young2018recent}. However, these solutions often suffer from incomplete language coverage or lack the capacity to understand more abstract, context-aware commands.

Recent advances in LLMs have sparked significant interest in applying generative language capabilities to driving tasks\cite{xu2025tell, yang2024llm4drive}. Although LLMs excel at parsing and generating complex human instructions, their direct deployment in real-time control is impeded by factors such as latency and safety constraints\cite{cui2024survey}. To address these issues, some researchers have proposed hybrid methods that couple high-level language understanding with downstream modules\cite{sha2023languagempc}. Despite these innovations, achieving a tight and reliable integration of LLM-based guidance with low-level driving controllers remains an open challenge. This gap underscores the need for architectures that can transform user instructions into feasible driving strategies while preserving responsiveness and safety.

\subsection{Hierarchical Decision Architectures}


In biology and cognitive science, fast-slow systems have been identified as a key organizing principle for intelligence, represented primarily by Kahneman's reactive System \uppercase\expandafter{\romannumeral1} and the deliberative System \uppercase\expandafter{\romannumeral2}\cite{kahneman2011thinking}. Inspired by these dual-process theories, researchers in robotics and artificial intelligence have explored hierarchical control architectures that separate high-frequency reactive control from low-frequency strategic planning\cite{botvinick2019reinforcement, bonnefon2020machine}.

In the context of autonomous vehicles (AVs), fast-slow paradigms have emerged to address the tension between quick reactive decisions, including collision avoidance, and more time-consuming deliberations including route planning and complex environment interpretation\cite{sha2023languagempc}. Recent works leverage learning-based methods for either the high-level or low-level layer, but rarely combine extensive user instructions with a genuinely human-oriented “slow” decision module\cite{mei2024continuously, fang2025interact}.

By drawing on established concepts from cognitive science and leveraging modern machine learning techniques, hierarchical fast-slow systems can capture both long-term goals and short-term safety requirements, ensuring that autonomous driving decisions respect diverse and evolving demands\cite{xu2024drivegpt4}.

\section{Problem Formulation}
We model our autonomous driving task as a Partially Observable Markov Decision Process (POMDP) denoted by the tuple \(\langle \mathcal{S}, \mathcal{A}, \mathcal{O}, P, R, \gamma \rangle\). For the ego vehicle may not have perfect information about distant or occluded vehicles, the agent receives an observation \(o \in \mathcal{O}\), which captures only the most relevant vehicles within certain range.

The environment transition function \(P(s_{t+1} \mid s_t, a_t)\) determines how the system evolves from state \(s_t\) to \(s_{t+1}\) given an action \(a_t\). Since the agent’s decision-making process is based on partial observations, we focus on designing a policy \(\pi(a_t \mid o_t)\) that maximizes expected returns:

\begin{equation}
    \max_{\pi} \;\; \mathbb{E}_{s_0 \sim \rho_0,\, a_t \sim \pi,\, s_{t+1} \sim P}\Biggl[\;\sum_{t=0}^{T} \gamma^t\, R(s_t, a_t)\Biggr]
\end{equation}
where \( \rho_0 \) is the initial state distribution, \(T\) is the horizon, \(\gamma \in (0,1)\) is the discount factor, and \(R(\cdot)\) represents the reward function.

\subsubsection{Action Space}
In this work, high-level action decisions are modeled as discrete commands, while a low-level PID controller executes the chosen command to regulate vehicle continuous control inputs. Let \(\mathcal{A}\) denote the set of available high-level actions:

\begin{small}
    \begin{equation}
    \mathcal{A} = \{slow down, cruise, speed up, turn left, turn right\}
    \end{equation}
\end{small}

By separating high-level decision making from the low-level PID-based controller, we can simplify policy learning while ensuring stable and interpretable control.

\subsubsection{Observation Space}
At each timestep \(t\), the agent receives an observation \(o_t\) consisting of the ego vehicle’s own state and the states of up to \(n\) vehicles in its vicinity. We represent this observation as a feature matrix \(\mathbf{O} \in \mathbb{R}^{n \times d}\), where each row corresponds to one vehicle. Each row of \(\mathbf{O}\) has a fixed dimensionality \(d\):

\begin{equation}
    d_i = [x_i, y_i, v_{x_i}, v_{y_i}, y_{des_i}]
\end{equation}
where \((x_i,\, y_i)\) and \((v_{x_i},\, v_{y_i})\) represent the spatial coordinates and velocity of \(vehicle_i\).

As we do not account for explicit prediction of lane changes of environment vehicles, \(y_{des_i}\) simply equals their current \(y\) position, while the ego vehicle's \(y_{des_i}\) is treated as user preference, which reflects which lane or lateral position the user desires.

\subsubsection{Reward Function}
To guide the agent’s behavior toward safety, efficiency, comfort, and user preference fulfillment, we define a reward function:

\begin{align}
    R(s_t, a_t) &= R_{\mathrm{safe}}(s_t, a_t) + R_{\mathrm{eff}}(s_t, a_t) \nonumber \\
                &\quad + R_{\mathrm{comfort}}(s_t, a_t) + R_{\mathrm{pref}}(s_t)
\end{align}
where the safety component, \(R_{\mathrm{safe}}(s_t, a_t)\), penalizes collisions; the efficiency component, \(R_{\mathrm{eff}}(s_t, a_t)\), encourages higher speeds; and the comfort component, \(R_{\mathrm{comfort}}(s_t, a_t)\), penalizes abrupt maneuvers. In addition, the user preference component, \(R_{\mathrm{pref}}(s_t)\), rewards alignment of the ego vehicle’s lateral position with the user-specified lane.

\begin{figure*}
    \centering
    \includegraphics[width=0.9\linewidth]{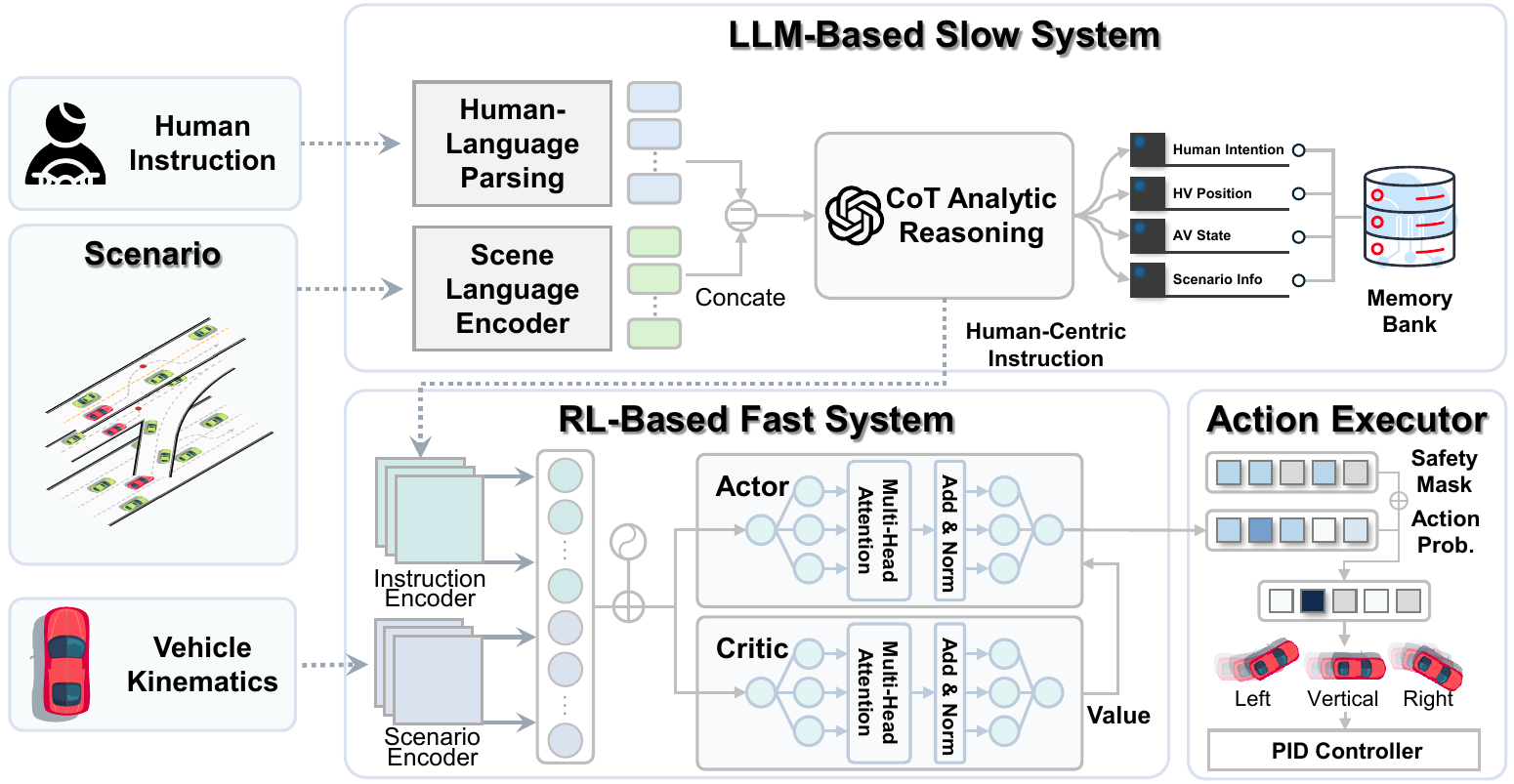}
    \caption{The proposed fast-slow architecture combines LLM guidance with RL. LLM interprets human intentions and converts them into structured instructions, while RL combines environmental information to make safe and robust decisions.}
    \label{fig:framework}
\end{figure*}

\section{Methodology}
\subsection{Framework overview}
The proposed system architecture, illustrated in Fig.~\ref{fig:framework}, is divided into two primary components: a slow system powered by LLM and a fast system driven by RL. In the slow system, the LLM combines human instructions with scene context to produce a structured directive reflecting user preferences. Concurrently, relevant scene data and decision events are stored in a memory bank, allowing the system to reference and refine its decision-making in future iterations.

The fast system integrates the observation space with LLM's human-centric instruction and inputs these into a policy network enhanced by an attention mechanism, enabling the agent to emphasize crucial elements of both the scene and user preferences. The low-level controller applies the selected action, which has been validated by a safety mask that filters out hazardous maneuvers.

\subsection{LLM-Based Slow System}
To parse and interpret high-level human instructions under varying traffic conditions, we introduce a slow system driven by LLM. To minimize spurious or unsafe recommendations, this system combines user commands, environmental context, and past driving experiences before producing structured directives for the fast system. 

Let \(I\) denote the raw user instruction, \(s_t\) the environment state at time \(t\), and \(\mathcal{D}\) a memory bank of historical scenarios and decisions. Concretely, the system performs the following steps:

\subsubsection{Scene Encoding} We define an encoding function \(\mathrm{Enc}(\cdot)\) that extracts relevant features from \(s_t\) and translates them into a concise textual description:

\begin{equation}
    E_t = \mathrm{Enc}(s_t)
\end{equation}
where \(E_t\) includes both static contextual information including road topology and dynamic elements including the positions and velocities of up to \(n\) nearest vehicles. By limiting the input to the most relevant neighbors, the mitigate potential hallucinations by the LLM can be reduced.

\subsubsection{Memory Retrieval} To incorporate prior experience, we retrieve historical data from \(\mathcal{D}\) using the cosine similarity function \(\mathrm{sim}(\cdot,\cdot)\) on sentence embeddings: 

\begin{equation}
    \mathbf{M}_{t} \;=\; \operatorname{arg\,max}_{\mathbf{M}_k \in \mathcal{D}} \mathrm{sim}\bigl(E(\mathbf{s}_t),\, E(\mathbf{M}_k)\bigr)
\end{equation}
where \(M_t\) represents a set of past scenarios resembling the current scene. These retrieved examples help guide the LLM by providing evidence of effective decisions in similar conditions.

\subsubsection{Instruction-Based Prompt Construction and Reasoning} A prompt construction function merges the user instruction \(\mathbf{I}_t\), the scene encoding \(\mathbf{E}_t\), and the memory retrieval \(\mathbf{M}_t\) into a composite prompt \(\mathbf{P}_t\): 

\begin{equation}
    \mathbf{P}_t \;=\; f(\mathbf{I}_t,\; \mathbf{E}_t,\; \mathbf{M}_t)
\end{equation}

To elicit step-by-step reasoning, the chain-of-Thought (CoT)\cite{wei2022chain} prompting strategy is employed, which instructs the LLM \(\Phi\) to explain intermediate logical steps explicitly.

\begin{equation}
    R_t \;=\; \Phi(\mathbf{P}_t)
\end{equation}
where \(R_t\) is the LLM’s textual response. This response is expected to account for both the environmental factors (\(\mathbf{s}_t\)) and user preference (\(\mathbf{I}_t\)), while also referencing any historical precedents retrieved from \(\mathbf{M}_t\). The goal is to ensure that each intermediate reasoning stage is checked for safety and consistency with user demands.

\subsubsection{Structured Directive Extraction} The last stage converts the LLM’s text-based response into a structured directive \(\hat{i}_t\) through a parsing function: 

\begin{equation}
    \hat{i}_t = \mathrm{Parse}(R_t)
\end{equation}

These parameters are then transmitted to the fast system (Section \ref{subsection_RL}), augmenting the agent’s observation with user-specific constraints. By separating free-form natural language from structured output, we ensure the compatibility with the RL policy network.

Overall, \(\hat{i}_t\) reflects a high-level human-oriented instruction that balances user preferences with real-time safety and operational constraints.

\subsection{RL-Based Fast System}
\label{subsection_RL}
While the slow system interprets high-level user instructions and encodes them as structured directives \(\hat{i}_t\), the fast system leverages a policy network based on multi-head attention mechanism to execute real-time control decisions. Let \(\pi_\theta(a_t \mid o_t, \hat{i}_t)\) denote the probabilistic policy, parameterized by \(\theta\), that outputs a discrete action \(a_t\) given the current observation \(o_t\) and user instruction \(\hat{i}_t\). To ensure robust convergence and stability, a policy optimization algorithm based on the actor-critic paradigm is adopted.

\subsubsection{Observation-Instruction Embedding}
During training, the observation space includes a randomized position for the target lane offset, \(y_\mathrm{dis}\), which will embed LLM-generated instruction \(\hat{i}_t\):

\begin{equation}
    o_t^i = (o_t^{env_i}, y_{dis}^i)
\end{equation}

where \(o_t^{env_i}\) denotes features including positions, velocities, and presence indicators for the closest \(n\) vehicles, and \(y_{dis}^i\) is randomized at initialization but subsequently aligned with \(\hat{i}_t\) during training, ensuring that the policy learns to interpret and act on user-oriented lane preferences within the same input embedding space.

\begin{figure}
    \centering
    \includegraphics[width=0.7\linewidth]{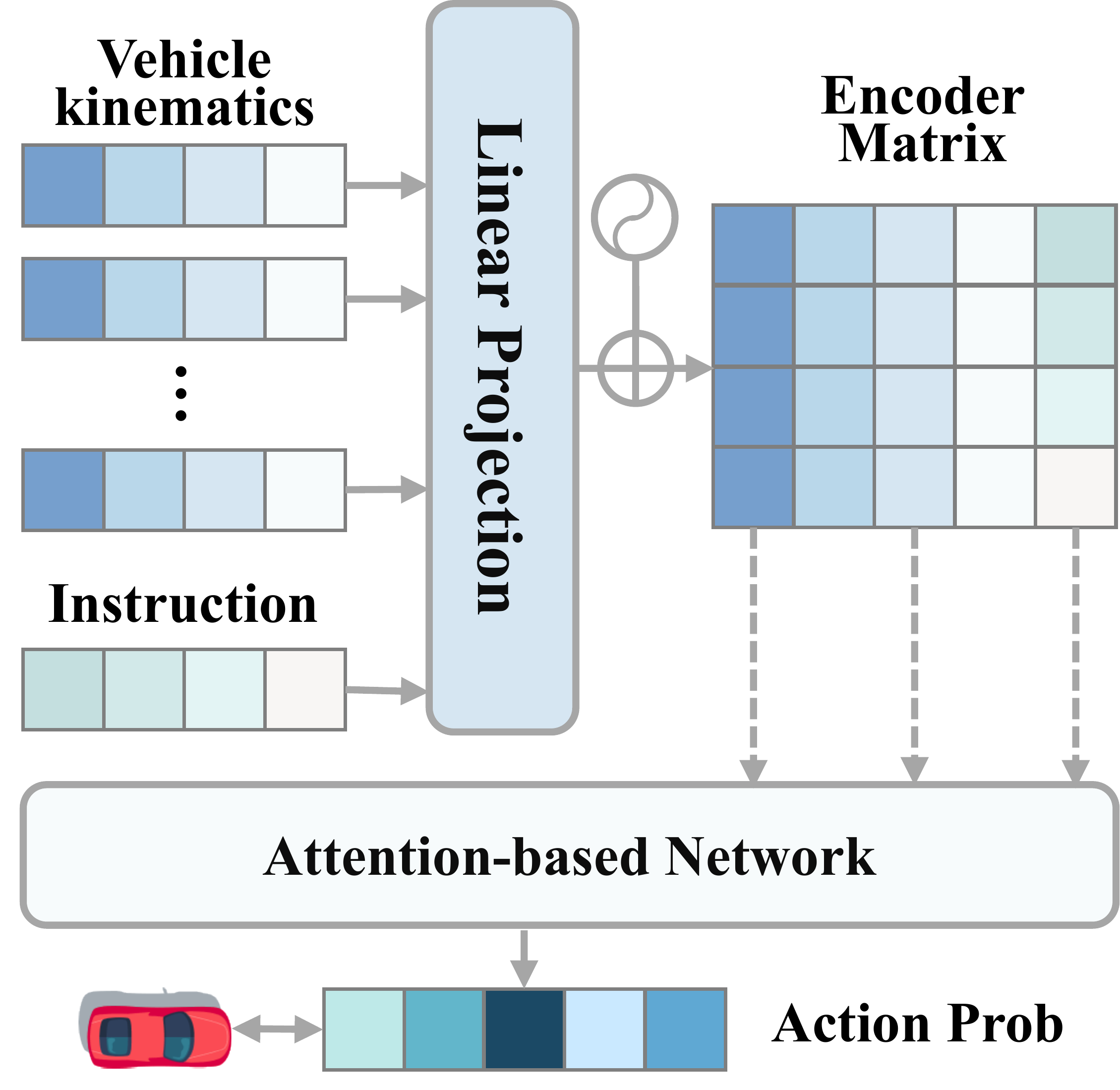}
    \caption{Overview of the embedding matrix of observation-instruction and the structure of the multi-head attention based policy network.}
    \label{fig:Network}
\end{figure}

\subsubsection{Multi-Head Attention Policy Network} To efficiently process the high-dimensional observation \(o_t\) and the embedded instruction \(\hat{i}_t\), we employ a multi-head attention mechanism in the policy network as shown in Fig.~\ref{fig:Network}. Let:

\begin{equation}
    z_t^{(0)} \;=\; f_{\mathrm{embed}}\bigl(o_t\bigr)
\end{equation}
represent the initial embedding of the observation vector, where \(f_{\mathrm{embed}}(\cdot)\) is a linear embedding layer. The subsequent encoder applies a series of multi-head self-attention (MHSA) and feed-forward layers (FFN)\cite{vaswani2017attention}:

\begin{equation}
    z_t^{(l+1)} \;=\; \mathrm{MHSA}\bigl(z_t^{(l)}\bigr) \;+\; \mathrm{FFN}\bigl(z_t^{(l)}\bigr)
\end{equation}
for layers \(l = 0,1,\ldots,L-1\), allowing the policy to focus attention on safety-critical features and user-lane alignment constraints in parallel. At the final layer \(l=L\), we obtain a context vector \(z_t^{(L)}\) that encapsulates environment information and user preference signals. This vector is then projected into a probability distribution over actions:

\begin{equation}
    \pi_\theta(a_t \mid o_t, \hat{i}_t) \;=\; \mathrm{softmax}\bigl(\;W_{\pi}\,z_t^{(L)} + b_{\pi}\bigr)
\end{equation}
where \(W_{\pi}\) and \(b_{\pi}\) are trainable parameters.

\subsubsection{Actor-Critic Optimization} The actor-critic approach simultaneously maintains a policy function \(\pi_\theta\) and a value function \(V_\phi\), parameterized by \(\theta\) and \(\phi\) respectively. The value function estimates the expected return from state \(s_t\) under policy \(\pi_\theta\):

\begin{equation}
    V_{\phi}(s_t) \;=\; \mathbb{E}\Bigl[\;\sum_{k=0}^{\infty}\gamma^k\,r_{t+k}\;\Big|\;s_t,\;\pi_\theta\Bigr]
\end{equation}
which serves as a baseline to reduce variance in policy gradient estimation. At each training iteration, we collect a batch of trajectories \(\tau = \{(s_t, o_t, \hat{i}_t, a_t, r_t)\}\), compute the advantage function:

\begin{equation}
    A_t \;=\; \sum_{k=0}^{K-1} \gamma^k\,r_{t+k} + \gamma^K\,V_\phi(s_{t+K}) - V_\phi(s_t)
\end{equation}
and update the policy parameters \(\theta\) via a gradient of the form:

\begin{equation}
    \nabla_{\theta}J(\theta) \;=\; 
\mathbb{E}_{\tau}\Bigl[\,\nabla_{\theta}\,\log \pi_\theta(a_t \mid o_t, \hat{i}_t)\,\bigl(A_t\bigr)\Bigr]
\end{equation}
subject to a constraint that limits excessive deviation from the old policy to improve stability. Meanwhile, the value function parameters \(\phi\) is updated to minimize the mean squared error:

\begin{equation}
    \min_{\phi}\;\mathcal{L}_{\mathrm{critic}} = \mathbb{E}_{\tau}\Bigl[\bigl(V_\phi(s_t) - V_{\mathrm{target}}(s_t)\bigr)^2\Bigr]
\end{equation}

\subsubsection{Safety-Based Action Executor}
At runtime, given the current observation \(o_t\) and the LLM directive \(\hat{i}_t\), the fast system infers a discrete action \(a_t\) from \(\pi_\theta\bigl(a_t \mid o_t, \hat{i}_t\bigr)\). To prevent potential collisions or other dangerous actions, the selected action passes through a safety mask that checks feasibility based on real-time distance, velocity, and lane occupancy constraints. If the action is validated, it is then executed by the vehicle’s low-level controller. 

\section{Simulation And Performance Evaluation}
\subsection{Experiment Setup}
\begin{figure}[htp]
    \centering
    \includegraphics[width=0.8\linewidth]{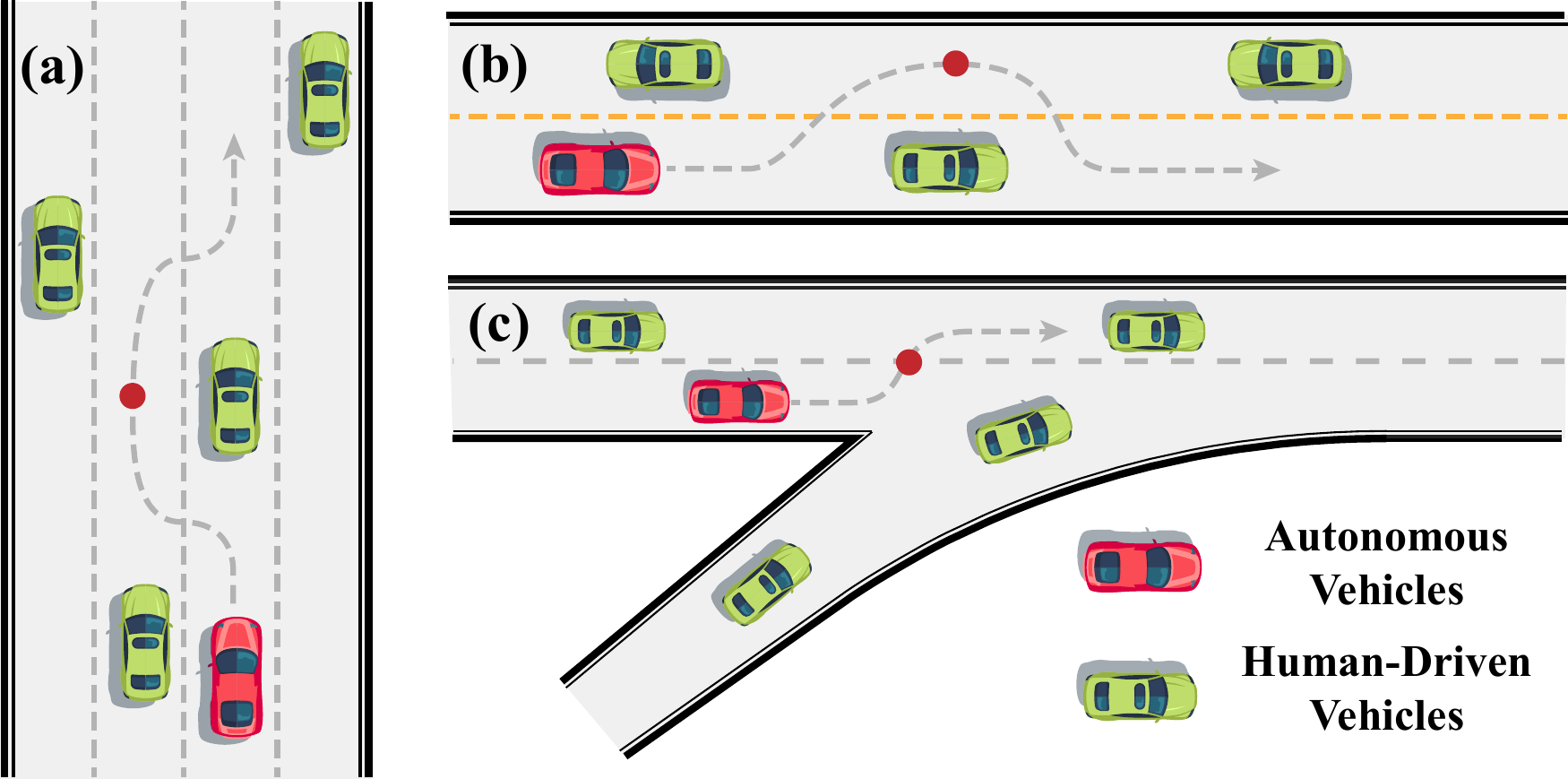}
    \caption{Three simulation scenarios designed to evaluate the framework’s capabilities from distinct perspectives: (a) high-speed driving, (b) right-lane vehicle merging, and (c) overtaking on a two-way road.}
    \label{fig:sce}
\end{figure}

We evaluate the proposed fast–slow architecture in a custom simulation stack that combines Highway-Env\cite{highway-env} and Gymnasium\cite{towers2024gymnasium}. As shown in Fig.~\ref{fig:sce}, three complementary scenarios are designed to probe different aspects of autonomous driving competence:  (a) a straight four‑lane highway that tests high‑speed cruising and multi‑lane planning, (b) a right‑side on‑ramp where the ego vehicle must negotiate cooperative merges, and (c) a two‑way rural road that requires safe overtaking of slower traffic in the presence of on‑coming vehicles. For each scenario we generate at least 100 episodes with randomized traffic seeds, arrival rates, and ego‑vehicle starting positions.

\subsection{Implementation Details}
The slow system uses GPT-4o-mini as the benchmark LLM for its fast reasoning and reliable logical reasoning capabilities. The fast system policy network adopts two attention heads and model dimension \(d_{\text{model}}=128\). The observation vector is augmented by an instruction slot \(y_{\text{dis}}\), randomly selected at episode start and overwritten at run‑time by the LLM‑derived lane preference \(\hat{i}_t\). The optimization of the policy follows an actor-critic scheme with an update of the trust region restricted by KL, using the discount factor \(\gamma=0.99\) and the learning rate \(5\times10^{-4}\). Each model is trained for \(10^{5}\) steps. All training and validation were performed on a computing platform equipped with an Intel(R) Core(TM) i7-14700K CPU, an NVIDIA GeForce RTX 4080 SUPER GPU and 32 GB of RAM.

\subsection{Performance Evaluation}

\subsubsection{Learning Efficiency and Convergence}
To assess the effectiveness of the proposed fast–slow architecture, we compare our model with three commonly used RL baselines: DQN\cite{mnih2013playing}, PPO\cite{schulman2017proximal} and A2C\cite{mnih2016asynchronous}. All agents are trained with identical parameters. 

\begin{figure*}[htp]
    \centering
    \includegraphics[width=\linewidth]{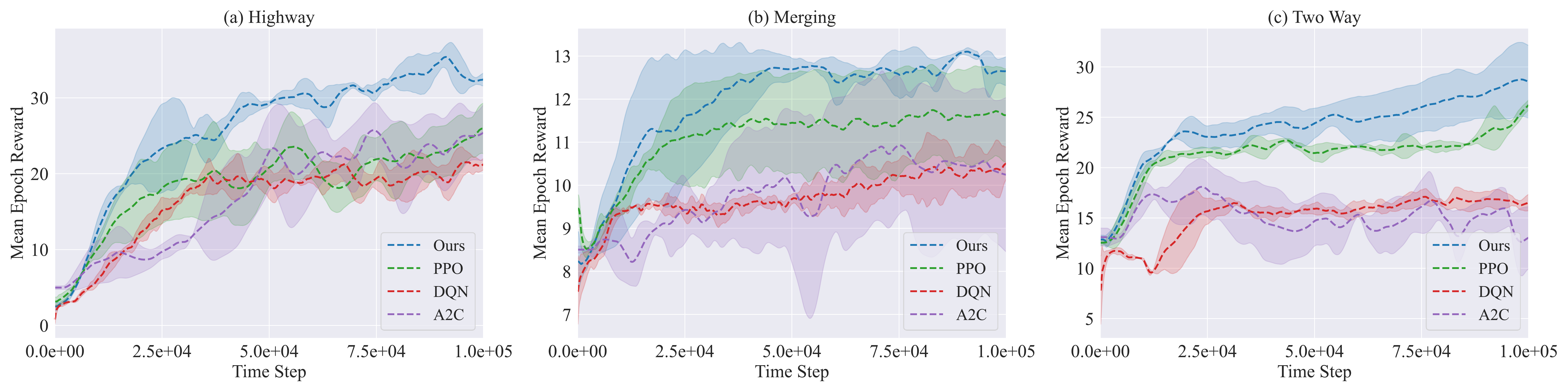}
    \caption{Comparison of reward function curves of the attention-enhanced Actor-Critic algorithm (Ours) and the existing state-of-the-art algorithm (PPO, DQN and A2C) during training in multiple scenarios, where the shaded area represents the 95\% confidence interval.}
    \label{fig:reward}
\end{figure*}

As shown in Fig.~\ref{fig:reward}, in every scenario the proposed agent climbs the return curve most rapidly and stabilizes at the highest asymptotic value. For example, on the four‑lane highway, our model surpasses PPO after roughly \(3\times10^{4}\) interactions and converges to a return 30\% higher than A2C. Similar advantages are visible in the merge and two‑way overtaking tasks. We attribute this sample efficiency gain to the multi‑head attention module that jointly attends to environmental features and the embedded directive, allowing the policy to generalize quickly across heterogeneous user commands.

\begin{figure}[htp]
    \centering
    \includegraphics[width=0.9\linewidth]{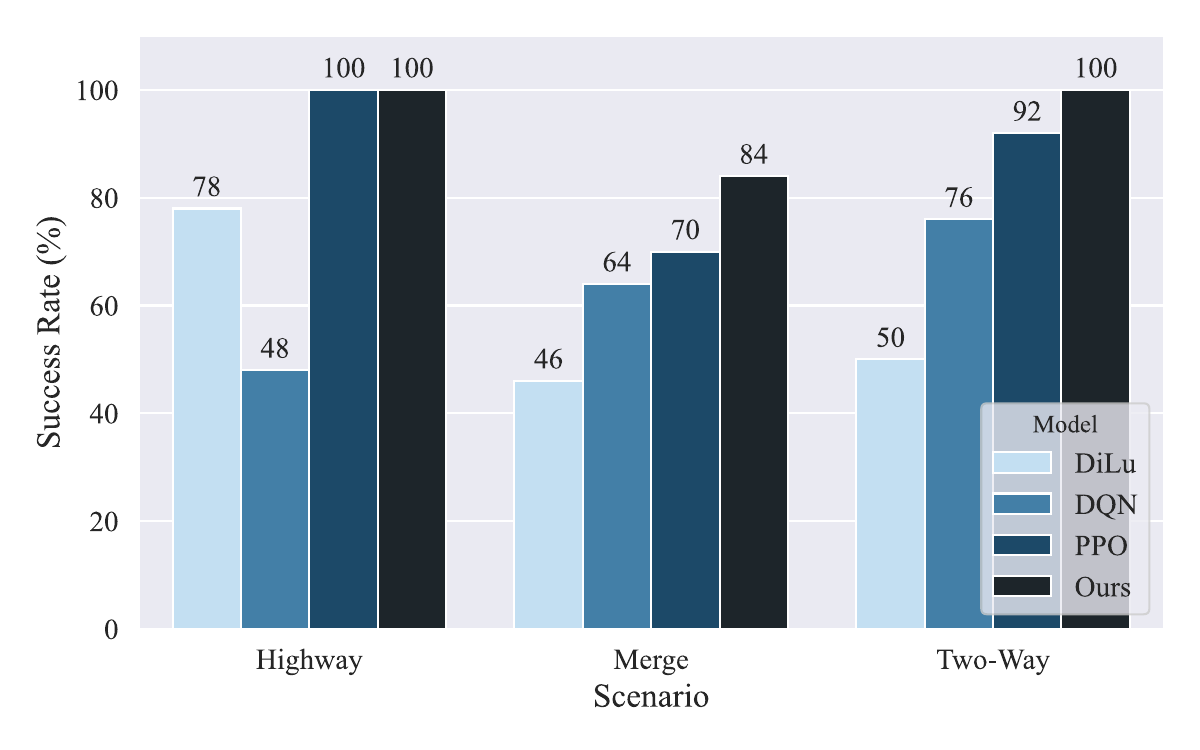}
    \caption{Comparison of the success rates of the proposed model and baselines model in three test scenarios.}
    \label{fig:comparison}
\end{figure}

\subsubsection{Behavioral Analysis}

For benchmarking, we adopt state‑of‑the‑art RL baselines: DQN for value‑based methods and PPO for policy‑based methods, together with the leading LLM algorithm, Dilu\cite{wen2023dilu}. As illustrated in Fig.~\ref{fig:comparison}, the proposed agent achieves the highest success rate in all scenarios. In addition, Table~\ref{Performance-Comparison} provides a more granular comparison using five interpretable metrics. Our agent offers the lowest acceleration variability simultaneously, which is an indicator of ride comfort. Nonetheless, the baseline outliers illustrate the cost of optimizing a single dimension.

PPO attains the largest min TTC on the highway by shadowing the slowest leader, satisfying safety but violating user requests for timely arrival. DQN pursues speed, yet its six‑fold increase in acceleration variance and collision‑prone merge behavior contradict both comfort and safety, reflecting an aggressive style that disregards comfort and occasionally causes collisions. DiLu interprets instructions well but lacks reactive finesse, halving its success rate in dense traffic and producing a stop‑and‑go speed trace. Unlike the baseline model, our proposed architecture realizes the intended lane or speed preference without sacrificing safety or comfort, fulfilling the core human-centric objective of instruction‑compliant, trustworthy autonomous driving.

\begin{table*}
\centering
\caption{The performance and aggregate indicators comparison of the proposed model and baselines in three scenarios}
\label{Performance-Comparison}
\begin{tabular}{llcccccc}
\toprule
\textbf{Sce} & \textbf{Model}  & \textbf{Average Speed(m/s)} & \textbf{Acceleration Variability} & \textbf{Min TTC(s)} & \textbf{Max Speed(m/s)} & \textbf{Min Speed(m/s)} \\
\midrule
\multirow{4}{*}{Highway} 
& DiLu & 23.30 & 0.43 & 23.92 & 26.11 & 22.00 \\
& DQN & 24.52 & 6.17 & 7.45  & 25.41 & 19.87 \\
& PPO & 22.47 & 0.41 & 31.88 & 25.35 & 22.00 \\
& \textbf{Ours} & \textbf{25.46} & \textbf{0.38} & 8.65  & \textbf{27.97} & 23.20 \\
\midrule
\multirow{4}{*}{Merging} 
& DiLu & 15.88 & 2.01 & 7.59  & 17.59 & 14.53 \\
& DQN & 18.69 & 3.72 & 2.62  & 19.98 & 14.12 \\
& PPO & 16.31 & 1.96 & 5.21  & 19.43 & 13.61 \\
& \textbf{Ours} & 14.51 & \textbf{0.32} & \textbf{18.93} & 16.69 & 13.19 \\
\midrule
\multirow{4}{*}{Two-Way} 
& DiLu & 7.55 & 1.65 & 1.47  & 7.99  & 5.07 \\
& DQN & 7.83 & 0.48 & 2.93  & 8.00  & 5.56 \\
& PPO & 5.05 & 0.30 & 10.36 & 7.01  & 3.21 \\
& \textbf{Ours} & \textbf{6.80} & \textbf{0.19} & 4.54  & \textbf{8.00}  & 4.02 \\
\bottomrule
\end{tabular}
\end{table*}

\subsection{Case Analysis}
To illustrate the qualitative advantages of the fast–slow architecture, we examine the two‑way overtaking scenario which is the most demanding of our test cases because it requires simultaneous reasoning about oncoming traffic, lane availability, and user intent. The complete experimental video can be accessed on our website\footnote{\href{https://drive.google.com/drive/folders/1K0WgRw1SdJL-JufvJNaTO1ES5SOuSj6p?usp=sharing}{Experimental Validation Video Weblink}} Conventional RL and LLM-based algorithms resolve this situation conservatively: they remain behind the slower lead vehicle for the entire episode, thereby maximizing safety at the cost of travel time. In contrast, our framework receives the user instruction “I’m in a hurry to get to work, I want to drive faster”. The slow system parses this abstract request, combines it with the current scene description, and emits a structured directive favoring a prompt but safe overtake.

\begin{figure}
    \centering
    \includegraphics[width=\linewidth]{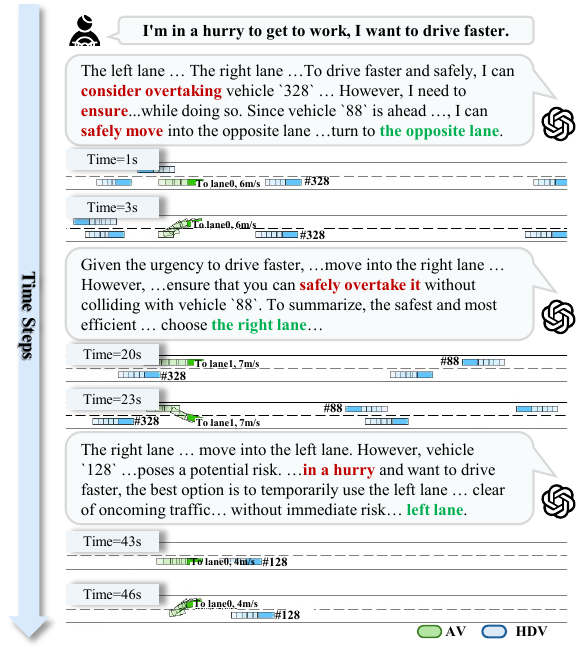}
    \caption{The specific performance of the proposed fast and slow system in a two-way overtaking scenario, where green represents AVs and blue represents HDVs.}
    \label{fig:case}
\end{figure}

As illustrated in Fig.~\ref{fig:case}, the slow system, after verifying that the oncoming lane is clear for at least the minimum overtake window, issues a left lane change command at \(t=1\,\text{s}\). The fast system validates the directive through its safety mask and completes the lane transition by \(t=3\,\text{s}\), accelerating to pass the slower vehicle. This behavior contrasts sharply with the baseline agents, which remain in the original lane and reduce speed.

The coordination between the two subsystems is not unilateral. At \(t=20\,\text{s}\), the slow system requests a return to the right lane to resume normal cruising. The fast system, however, detects that a faster vehicle is closing from behind; an immediate lane change would force the follower to brake sharply, risking a traffic flow shock. Consequently, the fast layer temporarily overrides the directive, first accelerating to widen the gap and then executing the lane change at \(t=23\,\text{s}\) when the manoeuvre can be completed without compromising the follower’s safety margin.

This case underscores three hallmarks of the proposed human‑centric design. First, the LLM‑based slow system successfully converts a high‑level natural language desire into a precise control objective. Second, the RL‑based fast system retains the autonomy to veto or delay instructions when real‑time safety constraints dictate. Third, the resulting behavior satisfies the user's intention for a faster journey while preserving comfort and safety, which can be hardly balanced by traditional models.

\section{Conclusion}
This study introduced a human‑centric decision‑making framework that couples an LLM-based slow system with an RL-based fast system. The LLM translates natural language instructions into structured directives, while the RL controller embeds these directives into its observation space and produces real‑time actions. Extensive simulations demonstrate that the proposed architecture attains the best balance among safety, efficiency, comfort, and command adherence.

Future research will extend the framework by enriching multi-modal inputs and outputs including voice tone and gesture cues to capture subtler aspects of passenger intent. Meanwhile, we plan to incorporate perception noise and partial observability to close the gap between simulation and real‑world deployment, followed by on-road trials. It is hoped that this work can explore a new feasible route for human-centric autonomous driving systems.

\ifCLASSOPTIONcaptionsoff
  \newpage
\fi

\bibliographystyle{unsrt}
\bibliography{related}

\end{document}